\documentclass{article}

\PassOptionsToPackage{numbers,compress}{natbib}
\usepackage[preprint]{neurips_2026}

\makeatletter
\renewcommand{\@noticestring}{}
\makeatother

\usepackage[utf8]{inputenc}
\usepackage[T1]{fontenc}
\usepackage{hyperref}
\usepackage{url}
\usepackage{booktabs}
\usepackage{amsfonts}
\usepackage{amsmath}
\usepackage{amssymb}
\usepackage{nicefrac}
\usepackage{microtype}
\usepackage{xcolor}
\usepackage{graphicx}
\usepackage{multirow}
\usepackage{array}
\usepackage{pifont}
\usepackage{xspace}

\newcommand{\cmark}{\ding{51}}
\newcommand{\xmark}{\ding{55}}
\newcommand{\wmark}{\ding{115}}

\newcommand{\NAME}{{DMI-Lib}\xspace}

\title{Enabling Performant and Flexible Model-Internal Observability for LLM Inference}

\author{%
  Nengneng Yu$^*$ \quad Sixian Xiong$^*$ \quad Yibo Zhao \quad Wei Wang \quad Zaoxing Liu \\\\
  Department of Computer Science \\
  University of Maryland, College Park
}

\begin{document}

\maketitle
\thispagestyle{plain}

\begin{abstract}
Today's inference-time workloads increasingly depend on timely access to a model's internal states. We present \NAME, a high-speed deep model inspector that treats internal observability as a first-class systems primitive, decoupling it from the inference hot path via an asynchronous observability substrate built from Ring$^2$, a GPU--CPU memory abstraction for capturing and staging tensors, and a policy-controlled host backend that exports them. \NAME enables the placement of observation points across a rich space of internal signals and diverse inference backends while preserving serving optimizations and adhering to tight GPU memory budgets. Our experiments demonstrate that \NAME incurs only 0.4\%--6.8\% overhead in offline batch inference and an average of 6\% in moderate online serving, reducing latency overhead by 2$\times$--15$\times$ compared to existing baselines with similar observability features. \NAME is open-sourced at https://github.com/ProjectDMX/DMI

\end{abstract}

\section{Introduction}
\label{sec:intro}

Large Language Models (LLMs) are rapidly becoming the core serving substrate for emerging AI-driven applications. As LLM-based deployments scale, {\em observability} of a model's internal execution data, beyond merely logging input prompts and output tokens, has become increasingly essential. A growing number of use cases have depended on timely access to the internal states when the model produces predictions, e.g., the longstanding pursuit of LLM interpretability~\cite{nanda2022transformerlens}, test-time alignment techniques that manipulate hidden states to steer model outputs~\cite{stolfo2024improving,du2025advancing}, and speculative decoding that leverages a target model's internal states to accelerate a draft model's inference~\cite{li2025eagle, dart, chen2026make}. Furthermore, activation probes can monitor high-stakes interactions at far lower cost than LLM-based monitors~\cite{mckenzie2025detecting}, and recent hallucination detectors exploit the cross-layer dynamics of hidden states rather than outputs alone~\cite{zhang2025icr}.

To enable these observability use cases, researchers have been retrofitting existing inference stacks with additional extraction logic. However, these efforts fall into two unsatisfying categories:
(1) {\em ad-hoc} observation built on extension mechanisms provided by PyTorch, which serves as the standard model implementation framework for many modern LLMs~\cite{paszke2019pytorch}; or (2) {\em inference-engine-bound} functionality, such as engine-provided APIs or output interfaces, whose access semantics are defined and implemented by the engine itself.

Neither approach provides a complete solution. Methods built on PyTorch extensions inherit the limitations of the underlying extension surface: for example, \texttt{register\_forward\_hook} typically exposes results only at module boundaries, while Python-facing extensions can introduce nontrivial overhead. Inference-engine-bound methods, in contrast, are tightly coupled to engine internals, so their accessibility is limited by what the engine explicitly exposes and has no portability across other platforms. More fundamentally, these approaches treat observability as an {\em auxiliary feature} coupled with execution, rather than as a systems concern with its own resource and performance constraints.

In this paper, we propose {\bf\em internal model observability as a first-class citizen}. We argue that the lack of adoption of real-time internal state access today is not a reflection of user demand: AI practitioners routinely need richer signals from their models, but rather a consequence of system infrastructure that was never designed to support it. Building a high-performance, real-time observability system, if successful, can lower the barrier for practitioners to understand and leverage internal insights from large models. Ideally, such an observability system should meet three key objectives:

\begin{itemize}
    \item {\em Broad observability coverage:} The system should expose a semantically rich and extensible space of internal vantage points (e.g., residual streams, attention patterns, MLP activations, KV-cache slices, and auxiliary control signals) across models and inference stacks, so that downstream users can express diverse analysis and control tasks without engine-specific modifications.
    \item {\em Low performance overhead:} The system should preserve the execution and memory contracts of high-performance model serving stacks, introducing only small, quantifiable slowdowns even under continuous monitoring, so that online workloads can safely enable internal-state access without violating latency or throughput targets.
    \item {\em High flexibility under hardware constraints:} The system should be able to adapt its data collection coverage and rate to heterogeneous hardware budgets (e.g., PCIe bandwidth, GPU memory headroom, storage throughput) via explicit policies, ensuring stable serving behavior while still extracting as much actionable internal signal as resource limits allow.
\end{itemize}

To meet the above requirements, we introduce {\bf \NAME}, a high-performance, model- and inference pipeline-agnostic {\bf D}eep {\bf M}odel {\bf I}nspector that systematically collects, transforms, and stores arbitrary internal states to enable downstream use cases. By decoupling observability from the inference hot path, \NAME not only enables existing use cases such as interpretability, test-time alignment, and speculative decoding, but also unlocks new ones that were previously impractical, such as real-time safety monitoring, where continuous extraction of internal states allows operators to detect malicious interactions and trigger early stopping before the generation completes~\cite{zou2023representation}.

In particular, \NAME introduces \texttt{HookPoint}, a lightweight, CUDA graph-compatible collection primitive that can be placed at arbitrary locations in PyTorch models to expose diverse internal states without modifying inference engine backbone or relying on engine-specific APIs, achieving broad observability coverage. To preserve serving performance, \NAME routes captured tensors through $Ring^2$, a GPU--CPU co-designed staging abstraction that keeps data movement inside CUDA graphs and a dedicated GPU-side ring buffer, then drains it asynchronously on the host via a data exporter, maintaining execution and memory contracts with only small, quantifiable overheads. Finally, to operate robustly under tight hardware budgets, \NAME combines reconfigurable hook with a runtime policy manager that enforces complete or best-effort export policies, adapting data rate and fidelity to interconnect and memory constraints while preserving serving latency and stability.

We integrate \NAME with Hugging Face~\cite{wolf-etal-2020-transformers} and vLLM~\cite{vllm} through an implementation consisting of 11K lines of CUDA, C++, and Python code. Our evaluation shows that \NAME incurs only 0.4--6.8\% overheads on offline batch inference and about 6\% on online serving, reducing latency by $2$-$15\times$ compared to various baselines with comparable observability.

The rest of this paper is organized as follows.
Section~\ref{sec:background} provides background on LLM inference and motivates the need for a dedicated observability system. Section~\ref{sec:challenge} to \ref{sec:design} identify challenges and present designs that address them. Section~\ref{sec:implementation} to \ref{sec:evaluation} discuss implementation and present a comprehensive evaluation. The paper concludes with discussions of related work in Section~\ref{sec:related} and conclusions in Section~\ref{sec:conclusion}.

\section{Background and Motivation}\label{sec:background}

In this section, we first describe the LLM inference and the need for a performant and flexible internal observability system. We then discuss existing tools and their limitations in enabling real-time monitoring capabilities.

\subsection{Transformer Model and Inference Infrastructure}

Modern LLMs are built on the Transformer architecture. In a standard dense, decode-only Transformer, the main components include an embedding layer, a stack of Transformer layers (each comprising self-attention and MLP modules), and residual connections with normalization layers~\cite{vaswani2017attention}. Recent models extend this dense design with architectural variants such as Mixture-of-Experts (MoE) layers~\cite{jiang2024mixtral,shazeer2017outrageously} and modified attention or sequence-processing blocks~\cite{dao2022flashattention,gu2024mamba}, further increasing architectural diversity.

Model architectures are commonly implemented in PyTorch~\cite{paszke2019pytorch}, which represents neural network components as \texttt{nn.Module} objects. Each \texttt{nn.Module} encodes a portion of the computation graph, such as an attention block, an MLP block, or an entire Transformer layer, and composing these modules defines both the structure of the model and the dataflow between stages. This executable-graph view enables developers to express complex model logic in a uniform abstraction while leaving low-level kernel selection and graph optimization to the framework.

LLM inference systems~\cite{vllm, zheng2024sglang, trtllm, tgi2023, llamacpp2023} provide the execution infrastructure for these model graphs. They are responsible for batching requests, scheduling token generation, managing KV-cache memory, organizing device buffers, and invoking optimized kernels to execute the computation described by the model. Modern serving systems such as vLLM~\cite{vllm}, SGLang~\cite{zheng2024sglang}, and Hugging Face TGI~\cite{tgi2023} decouple high-level model definitions from serving-time concerns: their backends implement scheduling, memory management, and kernel orchestration as a separate layer that can be reused across rapidly evolving model architectures with minimal integration effort.

\subsection{Example Downstream Applications}

As LLMs continue to see wider deployment, the demand for access to internal model states is growing rapidly. We discuss multiple motivating downstream applications of a high-performance model observability system like \NAME.

\paragraph{Mechanistic interpretability.}
For model explainability, mechanistic interpretability has shifted from describing input-output behavior to systematically analyzing residual streams, attention patterns, MLP activations, and cross-layer representations in order to uncover the internal mechanisms underlying model behavior~\cite{ferrando2405primer,rai2024practical}. Collecting and analyzing a large volume of internal states during model execution has become the cutting edge approach for model understanding.

\paragraph{Online model monitoring.}
For online monitoring and reliability, prompts and final outputs alone are no longer sufficient to capture risk signals. Recent work shows that activation probes~\cite{kramar2026building} can detect high-stakes interactions at much lower cost than LLM-based monitors~\cite{mckenzie2025detecting}. Meanwhile, some work uses cross-layer hidden-state dynamics for hallucination detection rather than relying only on final generations~\cite{zhang2025icr}. Internal states therefore become a direct signal source for online monitoring and reliability assessment.

\paragraph{Inference time control and alignment.}
Activation engineering~\cite{zou2023representation,turner2024steering,zhang2025precise} frames inference-time activation modification as a form of test-time control; and recent safety-alignment work ranks candidate responses directly using intermediate hidden states~\cite{du2025advancing}. Unlike prompt engineering or offline finetuning, these approaches place internal states inside the control logic of the inference path, making access to and manipulation of intermediate representations a new systems requirement.

\paragraph{Serving system optimization.}
Model internal states are increasingly used for serving optimization. EAGLE-3~\cite{li2025eagle} elevates speculative decoding to the feature level and exploits internal features to improve inference efficiency; subsequent work~\cite{dart} further uses exported hidden states to train draft models and improve draft acceptance rates.

\subsection{Existing Tools and Limitations}

In model serving, mainstream systems such as vLLM~\cite{vllm} and Triton inference server~\cite{triton_inference_server} have been focused on system- and hardware-level observability by exposing Prometheus-compatible system metrics~\cite{prometheus} (e.g., GPU and memory utilization), structured logs (e.g., logging errors and system operations), and distributed tracing. Popular tools such as Nsight Systems~\cite{nvidia_nsight_systems}, PyTorch Profiler~\cite{paszke2019pytorch}, and NCCL RAS~\cite{nccl} cover execution behavior at the levels of kernels, streams, memory, and communication. System- and hardware-level observability is widely deployed for performance or reliability optimizations~\cite{deng2025minder, wang2025reliable}. In contrast, our \NAME aims to collect the internal, intermediate data during model execution path.

Existing efforts to provide runtime observability of model internals remain sparse. Hugging Face~\cite{wolf-etal-2020-transformers} generation utilities can only return a subset of internals, including logits, attentions, and hidden states. vLLM and SGLang provide limited internal extraction or plugin mechanisms; TensorRT-LLM~\cite{trtllm} introduces debug tensors for inspection; and PyTorch hooks~\cite{paszke2019pytorch}, while flexible, are also positioned as offline debugging and profiling tools by official documents, indicating significant performance overheads at runtime. These interfaces are largely predefined model outputs, engine-specific debugging paths, or instrumentation intended for analysis prototypes, rather than a unified and general-purpose observability system for online serving.

\begin{figure}[t]
    \centering
    \includegraphics[width=0.55\linewidth]{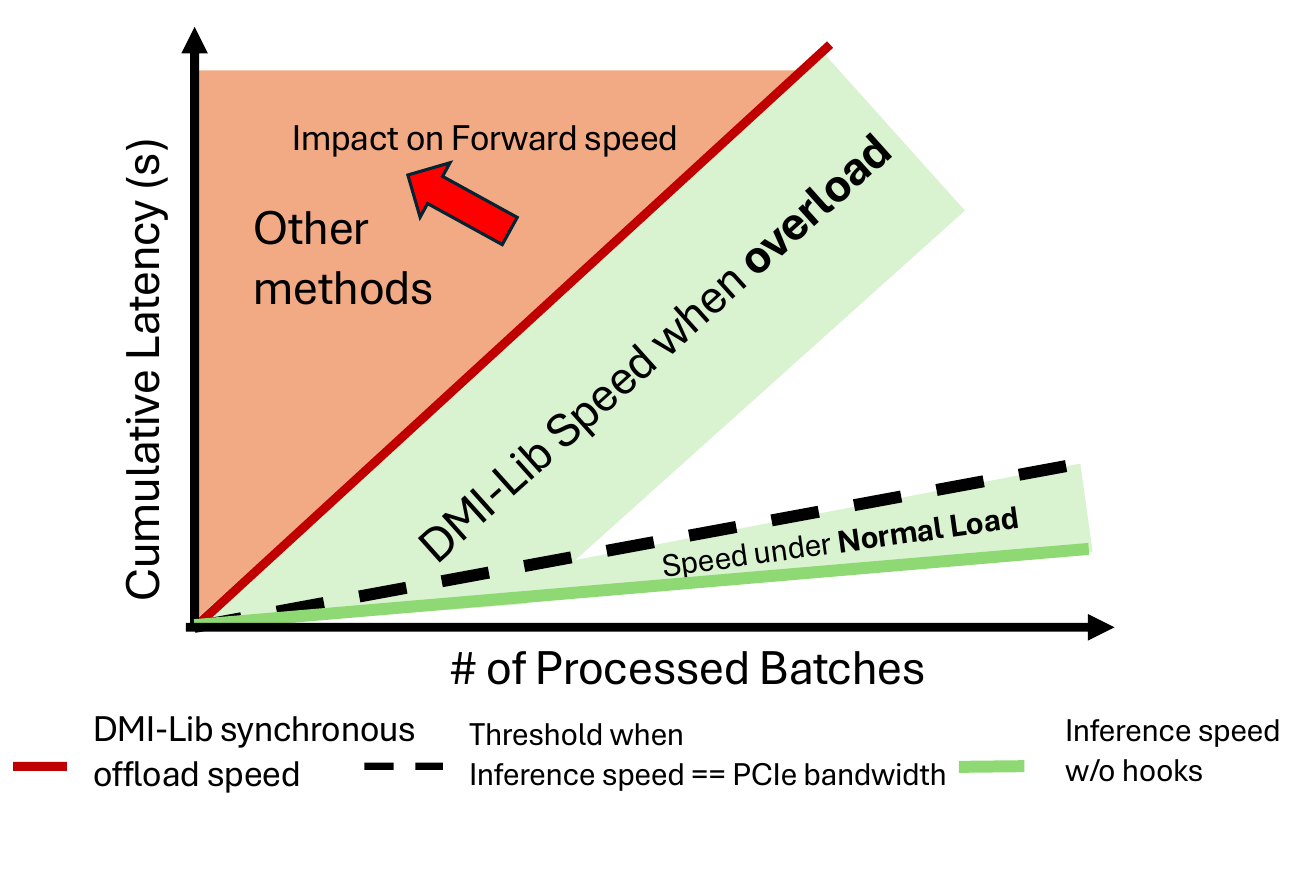}
    \caption{\textbf{Overview of \NAME vs.\ other methods:}
    When data speed stays below PCIe bandwidth, \NAME is close to the original inference speed. Once overload, the effective speed converges toward synchronous offloading.}
    \label{fig:overload}
\end{figure}

\paragraph{Summary.}
In parallel with the growing number of downstream applications, the right observability tool for model internals has largely been a missing piece. As summarized in Figure~\ref{fig:overload}, \NAME offers a solution that is both flexible and compatible with the constraints of high-performance inference.

\section{Challenges and Key Ideas}
\label{sec:challenge}

With the current limitations discussed above, we distill three concrete system challenges for online monitoring in high-performance LLM serving.

\paragraph{Challenge 1: Flexible observability over the entire inference path.}
\label{chal1}
LLMs expose many candidate vantage points, from hidden states and logits to Q/K/V projections and MLP activations, and operators differ across engines such as vLLM~\cite{vllm}, SGLang~\cite{zheng2024sglang}, and TensorRT-LLM~\cite{trtllm}. An observability system must allow users to target arbitrary points along this path while remaining portable across engines, without bespoke instrumentation.

\paragraph{Key idea.}
Rather than manually changing each inference engine to expose additional outputs, we treat data collection points as a general data-capture abstraction. With minimal effort, users can place collectors arbitrarily along the inference path, and \NAME then manages data capture and transport without relying on backend-specific APIs or implementations.

\begin{figure}[t]
    \centering
    \includegraphics[width=0.95\linewidth]{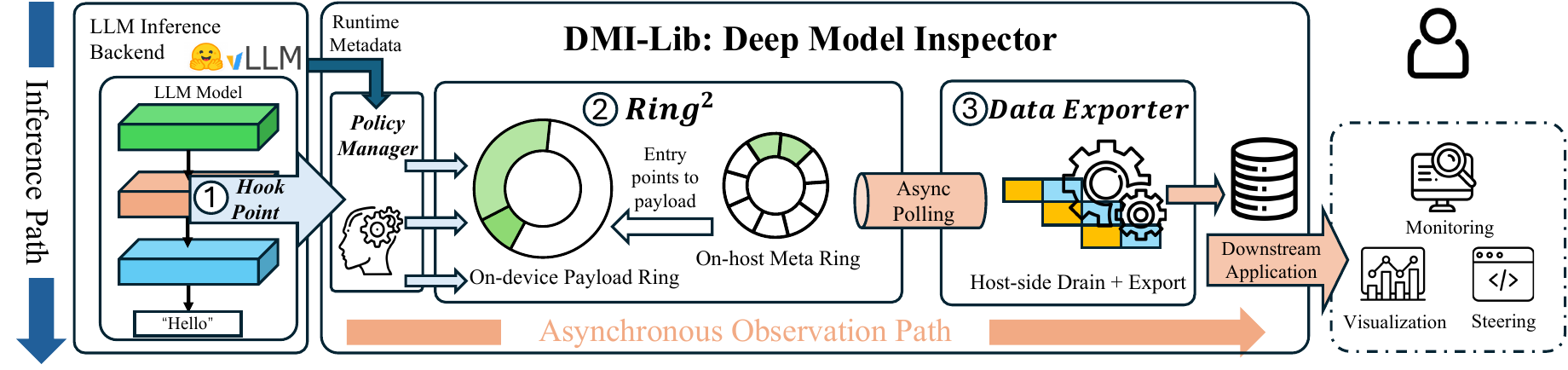}
    \caption{\textbf{DMI-Lib Overview.}}
    \label{fig:overview}
\end{figure}

\paragraph{Challenge 2: Collecting data without perturbing the on-device inference pipeline.}
\label{chal2}
Modern serving systems aggressively optimize GPU execution and memory: compiler-driven graph optimizations and CUDA Graph replay assume fixed traces and minimal host interaction, and GPU memory is tightly budgeted for KV cache~\cite{vllm} and batching~\cite{yu2022orca} with aggressive tensor reuse~\cite{paszke2019pytorch}. A naive data-collection pipeline breaks these assumptions, extending tensor lifetimes, invalidating graph capture and replay, and competing with KV cache and batching capacity. Observability must be decoupled from the inference fast path to preserve execution and memory guarantees.

\paragraph{Key idea.}
We introduce a separate observability data path for tensor capture and transport. By decoupling the main computation and observability paths, we preserve backend execution optimizations (e.g., compiler- or replay-based execution) and the serving memory contract, so observability minimally interferes with resources reserved for serving techniques such as KV cache and batching.

\paragraph{Challenge 3: Reconciling user requirements under hardware limits.}
\label{chal3}
The observability pipeline is a cross-device data link whose throughput is capped by interconnect bandwidth such as PCIe; when tensors are produced faster than they can be drained, backpressure builds on the GPU, inflating memory usage and eventually blocking computation. We envision that users will have diverse and sometimes conflicting goals: some prioritize data completeness, others prioritize strict latency isolation, and continuous monitoring may instead tolerate some kind of sampling. So no single static strategy can satisfy all scenarios within a fixed bandwidth budget.

\paragraph{Key idea.}
Rather than imposing a single static strategy, we manage backpressure at two levels: (1) Hook-level filtering lets users reduce observed data volume at the graph-topology level before data generation, and (2) a runtime policy manager resolves residual pressure per step---stalling for completeness or selectively dropping requests to keep inference unblocked, based on user-specified policies.

\section{\NAME Design}
\label{sec:design}

\NAME enables high-performance collection, analysis, and storage of internal states during the inference path. As depicted in Figure~\ref{fig:overview}, \NAME has three key components in its pipeline:
\begin{itemize}
    \item \textbf{(1) HookPoint} is a lightweight instrumentation primitive that is inserted directly into the model's forward graph so that when execution reaches it, it captures an arbitrary internal tensor and ready to hand it off.
    \item \textbf{(2) Ring$^2$} is the device-host co-designed memory staging mechanism that sits between on-device tensor capture and host-side processing, turning raw tensors into host-consumable records in an isolated buffer separate from the main inference path.
    \item \textbf{(3) Data Exporter} is the asynchronous host-side engine that continuously drains staged data from $Ring^2$, reconstructs tensors with their runtime metadata, and delivers them to memory and storage for downstream applications.
\end{itemize}

In addition to the data pipeline, \NAME has a runtime policy manager that configures which requests within each batch are observed, enabling request-level dropping under user requirements and bandwidth constraints.

\subsection{HookPoint}
\label{sec:hook}

\paragraph{Limitations of existing API.}
Existing mechanisms for capturing internal model states fall short along three axes: \textit{performance}, \textit{flexibility}, and \textit{portability}.
One approach is to modify the model's forward function to explicitly return selected intermediate tensors, as in Hugging Face's \textit{output\_hidden\_states} function~\cite{tgi2023}; however, this tightly couples observability to the model interface and does not scale to a larger number of data collection points.
A second approach is to use PyTorch's \textit{register\_forward\_hook} or \textit{register\_forward\_pre\_hook}, which are native callback mechanisms for observing the inputs or outputs of an \texttt{nn.Module} during execution; however, they are limited to module boundaries in a model and cannot reach arbitrary intermediate tensors within a module.
A third option is to rely on backend-specific tensor inspection features, such as the debug output API provided by TensorRT, but such mechanisms are tied to a particular inference engine and cannot be reused across frameworks.

\begin{figure}[t]
    \centering
    \includegraphics[width=0.6\linewidth]{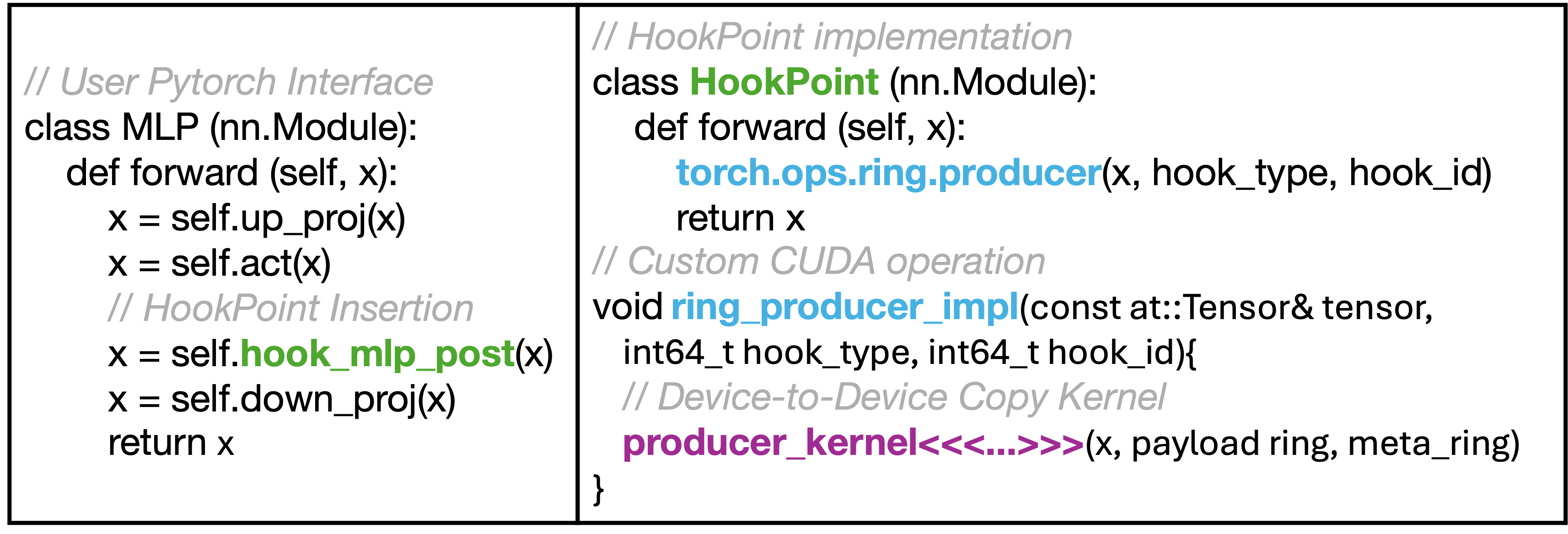}
    \caption{\textbf{HookPoint Execution Stack.}}
    \label{fig:pseudo_code}
\end{figure}

\paragraph{HookPoint: a two-layer instrumentation design.}
As shown in Figure~\ref{fig:pseudo_code}, at the \textit{Python interface layer}, HookPoint appears as a standard \texttt{nn.Module}, so it can be placed at any intermediate location in the computation graph without special framework support.
Adding a new observation point requires only two lines of code: one to insert the \texttt{HookPoint} at the target tensor location, and one in \texttt{\_\_init\_\_} to register it as a formal sub-module.
This design decouples the specification of observation points from any backend-specific API.

At the \textit{execution layer}, \texttt{HookPoint.forward()} dispatches the tensor through a custom PyTorch operator rather than a Python callback.
This operator launches a device-to-device (D2D) CUDA kernel that copies the target tensor into our $Ring^2$ staging structure (\S\ref{sec:ring2}) and returns immediately, leaving the original tensor unmodified on the main inference path.
This lets HookPoint expose arbitrary internal states through explicit graph-level instrumentation without relying on any backend-specific implementation or API.

\begin{table}[t]
  \centering
  \caption{Per-hook numeric impact on model output for a dense transformer layer (Qwen3-0.6B on vLLM).
    \textbf{\cmark}: bitwise-identical;
    \textbf{\wmark}: output differs due to kernel fusion disruption;
    \textbf{\xmark}: inaccessible due to FlashAttention/FlashInfer.}
  \label{tab:hook-impact}
  \renewcommand{\arraystretch}{0.9}
  \centering
  \begin{tabular}{llcc}
    \toprule
    \textbf{Category} & \textbf{Hook Point} & \textbf{bf16} & \textbf{fp32} \\
    \midrule
    Tokens & \texttt{token\_ids} & \cmark & \cmark \\
    \midrule
    \multirow{2}{*}{Residual Stream}
    & \texttt{resid\_pre/mid/final} & \cmark & \cmark \\
    & \texttt{hook\_ln1/ln2} & \cmark & \cmark \\
    \midrule
    \multirow{3}{*}{Attention}
    & \texttt{hook\_Q/K}          & \wmark & \cmark \\
    & \texttt{hook\_V/z/attn\_out}  & \cmark & \cmark \\
    & \texttt{attn\_scores, pattern} & \xmark$^{a}$ & \xmark$^{a}$ \\
    \midrule
    MLP & \texttt{mlp\_in/post/out} & \cmark & \cmark \\
    \midrule
    Output & \texttt{final\_ln, final\_logits} & \cmark & \cmark \\
    \bottomrule
  \end{tabular}
  \\[2pt]
  \raggedright\scriptsize
  $^{a}$ Fused attention techniques (FlashAttention), requiring additional kernel modifications.
\end{table}

The custom-operator design also helps addressing Challenge~2. Modern PyTorch-based model inference often relies on compiler-based execution optimizations, where the runtime transforms the original program into a more efficient execution graph. Python callback hooks interact poorly with such optimizations because they trigger synchronous host-side execution, and forces the engine to fall back to slower execution paths. In contrast, our capture logic is implemented as a native PyTorch operator that incurs no host-side synchronization, so it remains compatible with compiler-driven execution and brings minimal overhead during hook phase. Moreover, because the hook mechanism is expressed as a PyTorch-native operation, it can be reused across any inference frameworks that still use PyTorch model implementation.

\paragraph{Numerical effects of HookPoint instrumentation.}
Each HookPoint acts as both an execution barrier and a data materialization barrier: the captured tensor must physically exist in HBM in the model's storage dtype at the point where the hook fires. When a hook falls between operations that the compiler would otherwise optimize across, it can prevent the compiler from optimizing across the hook boundary, introducing small numeric differences such as FP32$\to$BF16$\to$FP32 precision round-trips. As shown in Table~\ref{tab:hook-impact}, the majority of hook placements produce {\em bitwise-identical} output to the unhooked model. A detailed analysis of the affected cases is provided in the supplementary material.

\begin{figure}[h]
    \centering
    \includegraphics[width=0.65\linewidth]{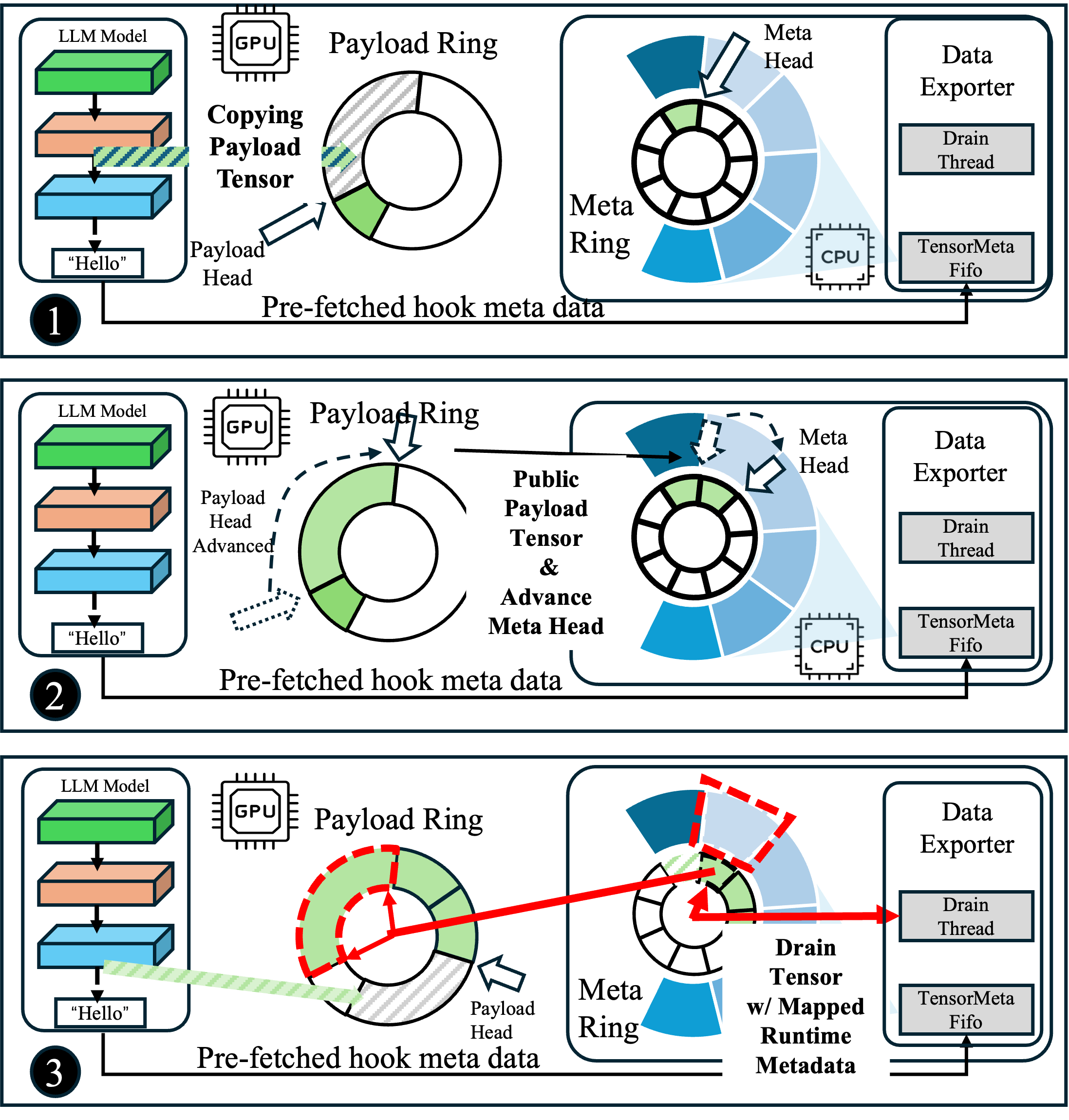}
    \caption{\textbf{$Ring^2$ Workflow.}}
    \label{fig:ring2}
\end{figure}

\subsection{$Ring^2$: GPU--CPU Memory Abstraction}
\label{sec:ring2}

After HookPoints are installed, \NAME requires a staging mechanism that can collect captured tensors on the GPU and expose them to the host without perturbing the inference fast path. $Ring^2$ provides this abstraction. It is a bounded, GPU--CPU split staging layer that transforms each captured tensor into a host-consumable record while remaining fully decoupled from the primary execution and memory paths of inference.

\paragraph{$Ring^2$ overview.}
$Ring^2$ is organized as two cooperating ring buffers: {\em payload ring} and {\em meta ring}, paired with a host-side FIFO queue {\em TensorMetaFIFO}. The payload ring stores raw tensor bytes and resides entirely in device memory, allocated via the operator \texttt{cudaMalloc}, so that writes remain on the GPU fast path. The meta ring stores fixed-size descriptors and is allocated in managed memory (\texttt{cudaMallocManaged}) with CPU-preferred placement, enabling low-cost host polling. The TensorMetaFIFO, maintained on the host, stores per-capture runtime metadata (e.g., request IDs, token ranges, shapes, dtypes) in FIFO order aligned with hook execution. TensorMetaFIFO is prepared before execution begins, since the serving runtime already knows which HookPoint will fire and its corresponding request/token context.

Figure~\ref{fig:ring2} illustrates the end-to-end workflow of $Ring^2$. The workflow proceeds in three steps: when forward execution reaches an active HookPoint, it invokes a custom CUDA operator that \textbf{(1)} copies the captured tensor into the payload ring using a block-parallel device-to-device transfer and advances \texttt{payload\_head}. \textbf{(2)} After the copy completes, the last finishing block publishes a descriptor into the meta ring at \texttt{meta\_head}. This descriptor encodes the payload offset, length, and a ready flag indicating availability for host consumption, and then advances \texttt{meta\_head}. \textbf{(3)} A dedicated drain thread continuously polls the meta ring. Upon observing a ready descriptor, it retrieves the corresponding payload region, matches it with the head of TensorMetaFIFO, and reconstructs a complete data record after asynchronous device-to-host transfer.

\paragraph{Benefits.}
$Ring^2$ organizes observation staging as a split ring buffer that keeps high-volume tensor payloads in device memory while exposing only compact descriptors in host-preferred memory, avoiding allocation overheads, memory and IO fragmentation associated with managing many disjoint fixed buffers (for tensors).

\paragraph{Memory placement rationale.}
The split design follows the asymmetric access patterns of the producer-consumer pipeline. The GPU-side device-to-device kernel launched by \texttt{HookPoint} is the producer and must write payload data at device bandwidth; thus, the payload ring is placed in device memory. The host-side D2H (device-to-host) drain thread is the consumer and repeatedly polls for readiness; thus, the meta ring is placed in CPU-preferred managed memory to minimize interconnect traffic and polling overhead. This mapping ensures that high-frequency data movement remains local to the GPU, while control-plane visibility remains efficient on the host, reducing the cost of frequent GPU--CPU coordination in the latency sensitive pipeline.

\paragraph{Decoupled metadata association.}
The split control path serves a second purpose: {\em it defers the pairing of a raw GPU tensor payload with its semantic metadata} (request ID, token range, hook name, etc.) until after the payload has exited the GPU hot path. The tensor capture path cannot tolerate synchronous host intervention without breaking the execution assumptions of modern optimized inference pipelines, which rely on static graphs, minimal CPU interaction, and aggressive kernel fusion. Instead, the D2D kernel records only the minimal offsets and lengths required for later reconstruction in the meta ring, while the richer runtime metadata is accumulated independently in the host-side TensorMetaFIFO. Once the payload has been drained to the host, \NAME reunites these two streams by matching ring entries with their FIFO counterparts, reconstructing a tensor with full semantic context entirely off the critical path. This decoupled association ensures that observation semantics remain precise without imposing synchronization or control-plane stalls on the main inference computation.

\subsection{Data Exporter}
\label{sec:backend}

After GPU-side publication and staging, \NAME relies on a host-side data exporter to complete the monitoring path. The backend continuously drains observation records from $Ring^2$ to the host, recovers their metadata and tensor structure, and delivers them to downstream storage and analytics tasks.

\paragraph{Data exporter workflow.}
A drain thread acts as the consumer of $Ring^2$ by continuously polling the meta ring. To avoid frequent stream-synchronization overhead from fragmented transfers, it adopts a batched strategy: once enough ready entries accumulate, or configurable occupancy, byte-count, or timeout thresholds are reached, it issues batched \texttt{cudaMemcpyAsync} operations to copy payloads from the GPU payload ring into a preallocated pinned-memory buffer on the host.

A pinned-to-page thread then consumes these payloads from pinned staging, copies them into pageable host memory, and immediately releases pinned buffer space. Although this introduces an extra host-side copy, it is necessary because pinned memory is limited. The data exporter then pairs payloads with runtime metadata (e.g., request id), reconstructs the tensor with the correct shape and dtype, and slices it by request and token range.

Finally, observability data slices, together with their metadata, can be sent to the persistence layer for offline storage and analysis. In current implementation, the data exporter connects to a ClickHouse sink storing internal data records.

\paragraph{Exporter optimization to avoid backpressure.}
We implement the data exporter as a bounded, multi-stage pipeline, decoupled via asynchronous threads and queues. The pipeline encompasses GPU-to-host DMA transfers, pinned-to-paged memory copies, tensor reconstruction/slicing, and final persistence. Each stage operates asynchronously to maximize throughput and overlap host-side processing with GPU execution. This design is critical because the efficiency of the data exporter directly determines the sustainability of the observation path under load. If host-side processing lags, backlogs accumulate in staging buffers and queues, propagating backward to the GPU-side observation path and eventually stalling the main compute stream.

\begin{figure}[t]
    \centering
    \includegraphics[width=0.55\linewidth]{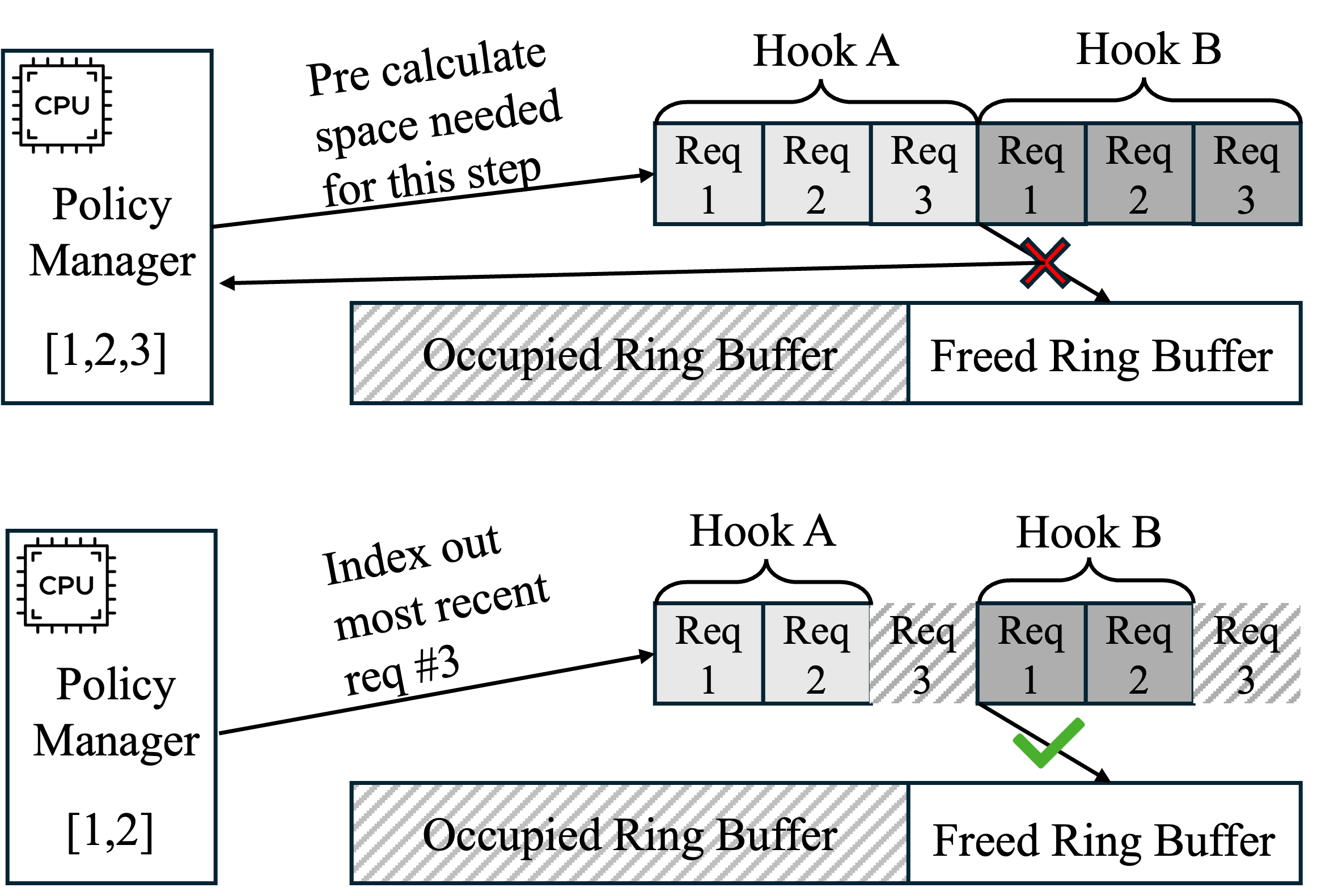}
    \caption{\textbf{Example of \textit{best-effort} runtime policy under the drop-recent strategy.}}
    \label{fig:request_dropping}
\end{figure}

\subsection{User Policies under Hardware Limits}
\label{sec:overload}

Fundamentally, the observation pipeline is constrained by cross-device bandwidth. Once the data collection stream persistently exceeds the sustainable export rate of the hardware interconnect, no system can preserve arbitrary observability without eventually paying either in memory growth or interference with the main inference path. However, a usable system needs to accommodate users' demands as much as possible, even under this hardware constraint. Some users prefer maximum observation data completeness and can tolerate slowdowns; others prioritize minimum interference with the inference path and would rather drop or sample observations under pressure. \NAME addresses this tension through a hierarchical overload-control design that maps limited bandwidth to user-specified policies.

Figure~\ref{fig:overload} illustrates the underlying execution model. When the observation generation rate is below the sustainable export bandwidth, \NAME remains close to the original inference speed with only modest overhead. Once generation outpaces export, buffered observations inevitably accumulate, and the system must eventually either stall the inference computation stream or reduce the observation load. Enlarging the ring can delay this transition by absorbing temporary bursts, but it cannot eliminate overload when the long-term generation rate exceeds the hardware constraints. \NAME, therefore, manages overload at two levels: it first reduces demand at the source through (1) hook filtering, and then applies a (2) runtime policy that resolves the remaining pressure according to the user's priority.

\paragraph{Hook filtering.}
The first layer of \NAME's overload control is hook filtering, which reduces data collection demand before any data is generated. Before a generation run begins, the user selects which hook set to enable; for example, collecting only attention activations, only MLP activations, or a custom subset. The HookPoints of disabled hooks degenerate to identity function in the forward path, so their corresponding D2D kernels do not appear in the execution graph. Hook filtering therefore operates at the graph-topology level, rather than discarding data after capture. Because this mechanism operates at hook granularity, it provides the most direct and lowest-overhead first layer of overload control.

\paragraph{Runtime policies.}
After hook filtering determines the set of active hooks, \NAME operates under user-defined policies. We present two simple policies below.

The {\bf\em completeness policy} is a default policy that \NAME collects and exports every observation data as defined by hook filtering from serving every request. This policy prioritizes observation completeness at the cost of potential inference slowdown under sustained overload.

The {\bf\em best-effort policy} attempts not to stall inference. Instead, when the ring buffer is under pressure, \NAME drops selective requests from observation rather than blocking the computation stream. \NAME provides two built-in dropping strategies: (1)~{\em Drop-recent.} As shown in Figure~\ref{fig:request_dropping}, the most recently added requests are dropped in the batch, so that \NAME retains observability on requests that are most likely to finish soon. (2)~{\em Keep-by-pattern.} The user specifies a predicate based on request content or request identifiers (e.g., requests whose prompt starts with a particular prefix, or requests with specific IDs). Requests matching the predicate are prioritized; when capacity is insufficient, other requests are dropped first.

\paragraph{Policy manager.}
Both policies are implemented by a host-side policy manager that runs before each forward step, using runtime metadata from the inference engine and the TensorMetaFIFO. Under the completeness policy, when the ring buffer is under pressure, the policy manager triggers a \texttt{flush} through the drain thread before proceeding with the forward step, directly relieving the ring pressure. Under the best-effort policy, the policy manager applies the user-selected dropping strategy and writes a batch-aligned keep/drop index vector into managed memory. The D2D kernel reads this vector at runtime and performs a gather-compact copy along the batched dimension, writing only the slices corresponding to retained requests into $Ring^2$. Because the index vector does not change the captured graph topology, updating it between steps remains fully compatible with CUDA Graph replay.

This policy manager gives users explicit control over the bandwidth tradeoff. Users who require completeness use the default policy. Users who prioritize latency isolation switch to best-effort with a dropping strategy: drop-recent to maximize finished traces under continuous monitoring, or keep-by-pattern for observation of specific requests.

\subsection{Support for Distributed Inference}
\label{sec:distributed}

Large models are routinely partitioned across multiple GPUs using parallelism strategies such as tensor parallelism (TP)~\cite{megatronlm}, pipeline parallelism~\cite{narayanan2019pipedream,huang2019gpipe}, and expert parallelism~\cite{lepikhin2020gshard}. \NAME supports distributed inference without cross-device coordination into the observation path.

Each GPU in a distributed deployment runs its own \NAME instance: a private $Ring^2$, a private drain thread, and a private host-side pipeline. No observation-related communication occurs between ranks within a step. Each rank captures whatever tensors are local to its device and tags every observation record with its parallelism coordinates. Downstream consumers can reconstruct full tensors by joining records across ranks using these tags together with the shared request and token identifiers. When the model is partitioned by layer, the layer index in each record identifies which pipeline stage produced it. When the model is partitioned along the hidden dimension, the distributed rank identifies which partition of the tensor the record contains.

\section{Implementation}
\label{sec:implementation}

\NAME is implemented in 6,600 lines of CUDA/C++ and 3,900 lines of Python. The C++ core comprises the $Ring^2$ data structures, D2D kernel, drain and pinned-to-page threads, and a database connector (ClickHouse sink~\cite{clickhouse}). The Python layer handles framework integration, hook installation, hook selection, and overload policy. GPU-side data-path components, including ring management and D2D copy, are implemented in CUDA.

\paragraph{Device-to-device kernel.}
The D2D kernel copies tensor data from the computation buffer into the $Ring^2$ payload ring on the GPU. It uses a size-tiered grid of 1--64 thread blocks (256 threads each), copying \texttt{uint4} (128-bit) elements in a grid-stride loop with tail bytes handled by low-numbered threads. After the copy, the last retiring block writes a 64-byte metadata descriptor into a CPU-preferred managed memory buffer, issues \texttt{\_\_threadfence()} for global visibility, and signals the drain thread by setting \texttt{ready\_seq} via a volatile store. The drain thread polls \texttt{ready\_seq} with an atomic acquire load; on consumption, it resets the slot to a sentinel, releasing it for reuse.

\paragraph{Framework integration.}
\NAME integrates with serving frameworks via lightweight wrapping, leaving backbone code unmodified. For vLLM, \NAME subclasses \texttt{Worker}, overriding \texttt{init\_device} to initialize the ring engine, \texttt{load\_model} to install HookPoints before CUDA Graph capture, and \texttt{execute\_model} to invoke \texttt{prepare\_step} prior to each forward pass. For Hugging Face, \NAME wraps \texttt{prepare\_inputs\_for\_generation}. In both cases, HookPoints are inserted at the desired observation sites and captured in model execution. Models declare their observation sites via a \texttt{HookSpec} interface, which provides the transport layer with the metadata template needed to pre-populate \texttt{TensorMetaFIFO} before each generation run. \NAME currently supports Hugging Face Transformers~\cite{wolf-etal-2020-transformers} and vLLM~\cite{vllm} as inference backends across a range of open models including Qwen3 and Llama series~\cite{qwen3,llama3}.

\section{Evaluation}
\label{sec:evaluation}

In this section, we evaluate \NAME's end-to-end performance under various workloads and validate the effectiveness of its core design choices. Our evaluation demonstrates:
\begin{itemize}
    \item \NAME incurs 0.4--6.8\% overhead on offline batch inference and $\sim$6\% on online serving, reducing latency overhead by 2$\times$ to 15$\times$ compared to baselines. (\S\ref{evl:offline}, \S\ref{evl:online})
    \item By enabling runtime policy, \NAME degrades gracefully under extreme PCIe bandwidth saturation, reducing end-to-end overload penalties by up to 12.4$\times$. (\S\ref{evl:overload})
    \item $Ring^2$ and asynchronous pipeline preserve the optimized compute path with only 1.36\% prefill and 5.05\% decode overhead; and the isolated memory design enables up to 98\% of the baseline maximum batch size. (\S\ref{evl:micro})
\end{itemize}

\subsection{Experimental Setup}

\paragraph{Model and server configurations.}
We use Qwen3-4B and Qwen3-14B~\cite{qwen3} models and Llama-3.1-8B~\cite{llama3} for end-to-end experiments. We use Qwen3-1.7B for our microbenchmark.
Our testbed consists of a compute node from an HPC center, equipped with an NVIDIA H100 80GB GPU and an Intel Xeon Platinum 8468 CPU, and a local GPU workstation for microbenchmarking, equipped with multiple NVIDIA GeForce RTX 4090 GPUs (24GB) and an AMD Ryzen Threadripper PRO 5955WX CPU.

\paragraph{Datasets.}
We evaluate on ShareGPT~\cite{sharegpt} and WildChat-1M~\cite{zhao2024wildchat}. Their average prompt/output lengths are 211/549 and 246.9/400.2 tokens, respectively.

\paragraph{Baselines.}
We compare \NAME against several baselines that support internal-tensor extraction.
\begin{itemize}
    \item \textbf{Python Callback with Flexible Hook Points.} This class provides flexible observation sites and can access intermediate tensors:
(1) \textbf{Torch Hooks} in PyTorch provide \texttt{register\_forward\_pre\_hook} and \texttt{register\_forward\_hook}, which return input/output from a module's \texttt{forward()} call, respectively. (2) \textbf{NNsight}~\cite{fiottokaufman2024nnsightndifdemocratizingaccess}, a Python library provides tracing-style inspection of foundation-model internals.
    \item \textbf{Serving Engine-Specific Tools.}
This class includes APIs and plugin-based tools tightly coupled to a serving engine.
(1) \textbf{HF Built-in Extraction.} Hugging Face \texttt{generate()} can return per-step \texttt{hidden\_states}, \texttt{attentions}, and \texttt{scores} (logits) through corresponding output flags. (2) \textbf{HF Stepwise Extraction.} This baseline uses Hugging Face model outputs for hidden states and logits, but replaces \texttt{generate()} with an explicit prefill/decode loop and attempts to compile the forward path.
(3) \textbf{vLLM Hook.} vLLM Hook~\cite{ko2026vllm} is a plugin for programming model internals on vLLM. We use it as an engine-coupled baseline for accessing internal states in a production-oriented serving stack. (4) \textbf{TRT-LLM (Debug API).} TensorRT-LLM~\cite{trtllm} is NVIDIA's inference engine providing a debug output feature that performs functionally similar internal state extraction.
\end{itemize}

\begin{table}[t]
\centering
\caption{\textbf{Observability coverage of offline inference baselines.} Column groups correspond to the two evaluation settings: \emph{limited hooks} (predefined engine outputs) and \emph{custom hooks} (user-defined internal states).}
\label{tab:baseline_hooks}
\begin{tabular}{l cc ccc}
\toprule
& \multicolumn{2}{c}{\textbf{Limited Hooks}} & \multicolumn{3}{c}{\textbf{Custom Hooks}} \\
\cmidrule(lr){2-3} \cmidrule(lr){4-6}
\textbf{Method} & \textbf{Hidden States} & \textbf{Logits} & \textbf{Q/K/V} & \textbf{MLP} & \textbf{Resid.} \\
\midrule
HF Built-in Extraction()    & \cmark & \cmark & \xmark & \xmark & \xmark \\
HF Stepwise Extraction   & \cmark & \cmark & \xmark & \xmark & \xmark \\
PyTorch Hooks    & \cmark & \cmark & \cmark & \cmark & \cmark \\
NNsight          & \cmark & \cmark & \cmark & \cmark & \cmark \\
\textbf{\NAME}     & \cmark & \cmark & \cmark & \cmark & \cmark \\
\bottomrule
\end{tabular}
\end{table}

\begin{figure}[t]
    \centering
    \includegraphics[width=\linewidth]{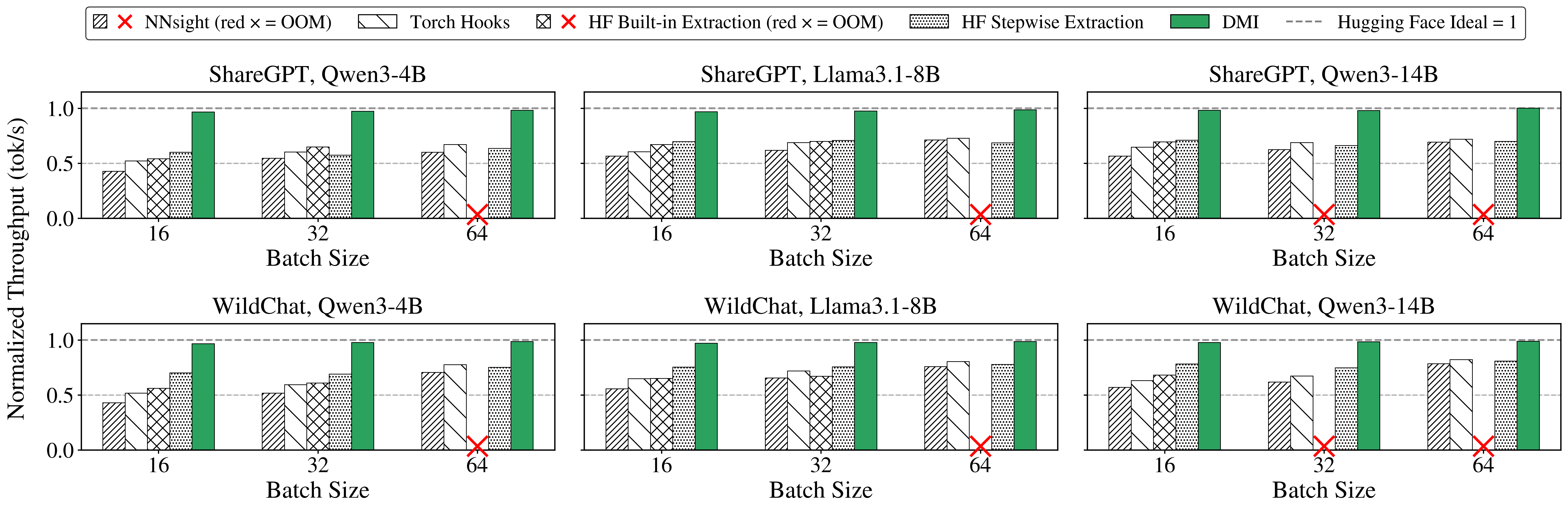}
    \caption{\textbf{Offline Performance with Limited Hooks:} 1 hidden-state hook per layer + 2 global hooks (final\_ln and logits), for a total of 38/34/42 hooks on Qwen3-4B, Llama3.1-8B, and Qwen3-14B, respectively.}
    \label{fig:offline_hs}
\end{figure}

\begin{figure}[t]
    \centering
    \includegraphics[width=\linewidth]{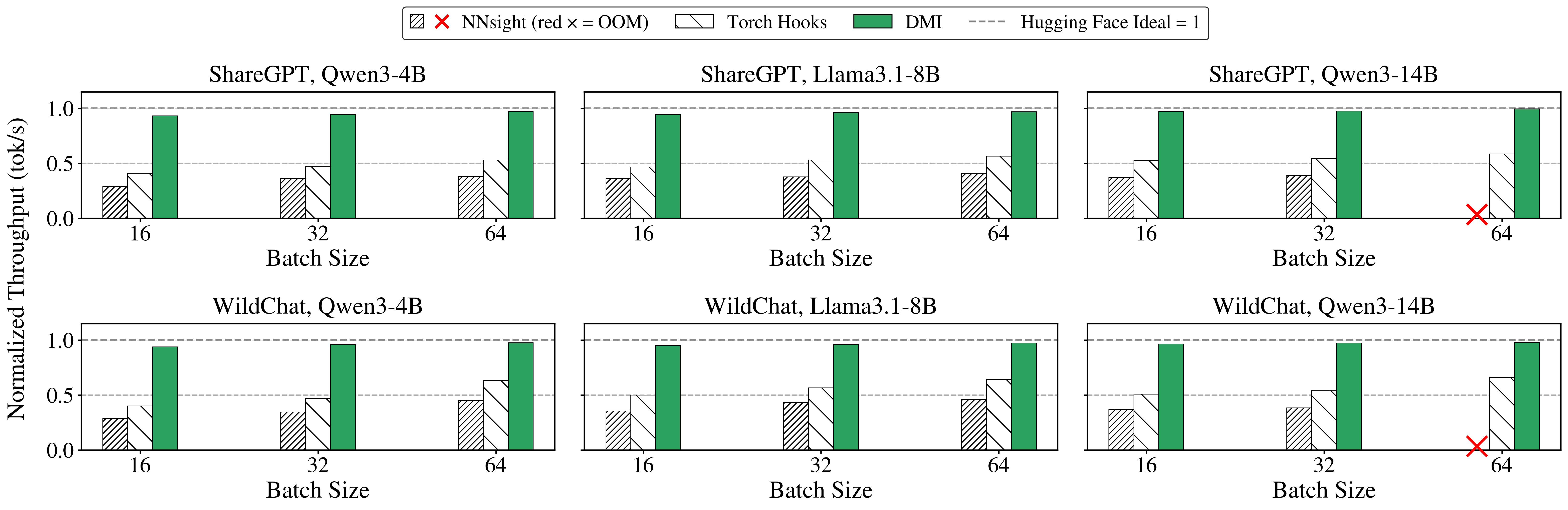}
    \caption{\textbf{Offline Performance with Custom Hooks:} 7 hooks per layer, for a total of 252/224/280 hooks on Qwen3-4B, Llama3.1-8B, and Qwen3-14B, respectively.}
    \label{fig:offline_hooks}
\end{figure}

We evaluate \NAME's end-to-end performance for both offline batch inference and online serving workloads, using Hugging Face as the offline inference backend and vLLM as the online serving backend. Our goal is to measure the overhead of enabling observability while preserving access to internal states, so we configure all baselines to extract internal tensors to host memory and deploy \NAME's ClickHouse~\cite{clickhouse} database on a tmpfs in-memory file system for consistency and fairness.

\subsection{Offline Batch Inference Performance}\label{evl:offline}

We begin with offline inference on an H100-80GB GPU, where the task is to run a fed dataset at once, and generate outputs end to end. We sampled 500 requests from both datasets. As summarized in Table~\ref{tab:baseline_hooks}, the baselines differ substantially in observability coverage: some methods are limited to predefined engine outputs, while others provide more flexible hook placements. To reflect this difference, we evaluate two settings separately: {\bf\em limited hooks}, where all methods observe equivalent internal states, and {\bf\em custom hooks}, where flexible user-defined hooks are required.

\paragraph{Limited hooks.}
We first compare methods on a limited set of predefined observation points. The results are shown in Figure~\ref{fig:offline_hs}. Across models, batch sizes, and datasets, \NAME incurs only 1.0--3.3\% overhead over running vanilla Hugging Face (ideal), with an average overhead of 1.9\%. This is substantially lower than all baselines. The strongest baseline is \textit{HF Stepwise Extraction}, which incurs 20.0--42.5\% overheads, with an average of 29.1\%. Although this approach performs better than other baselines, it requires manually rewriting the generation loop and remains limited to the predefined internal states exposed by Hugging Face. The slowest baseline is NNsight, which incurs 21.4--57.2\% overheads, with an average of 39.1\%.
We also observe that the naive \textit{HF Built-in Extraction}, which offloads internal states only after completing a full batch, runs into OOM at batch size of 64.

\begin{figure}[t]
    \centering
    \includegraphics[width=0.75\linewidth]{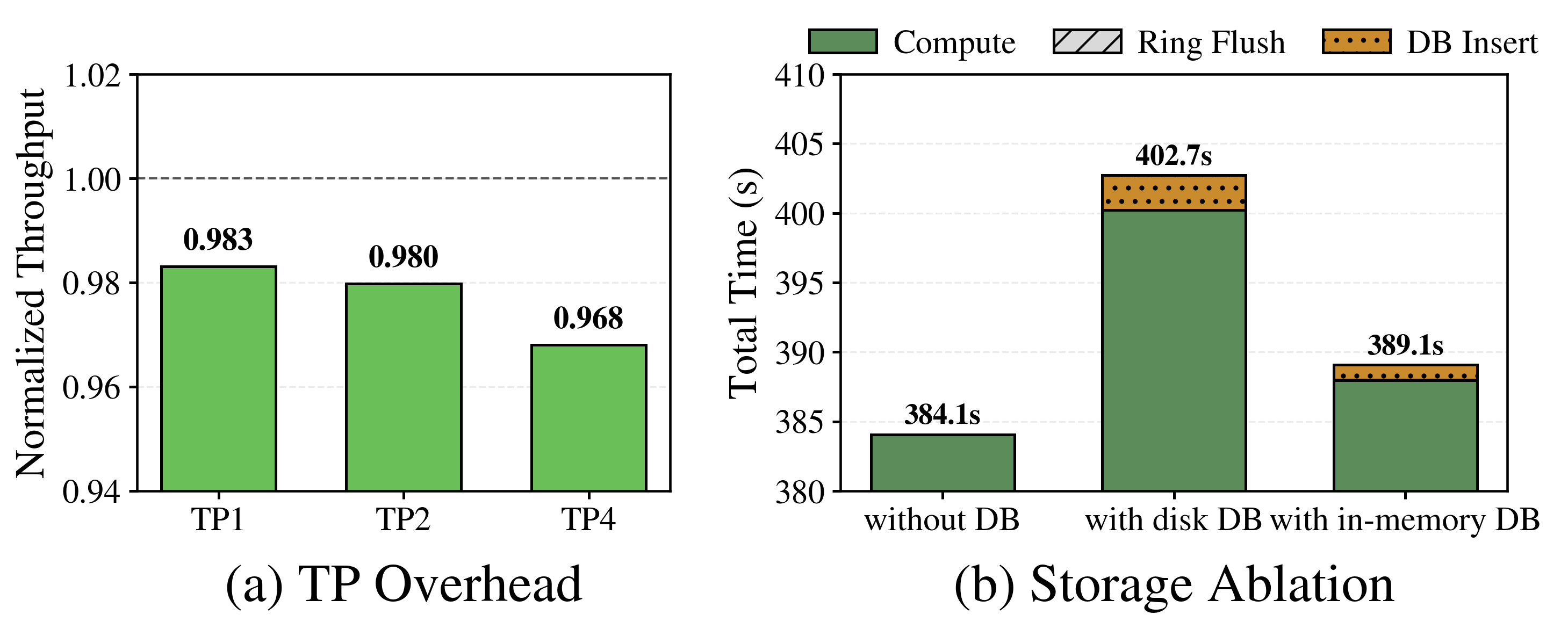}
    \caption{\textbf{Offline performance:} (a) Tensor-parallel performance. (b) Storage ablation study.}
    \label{fig:tpandstorage}
\end{figure}

\begin{figure}[t]
    \centering
    \includegraphics[width=\linewidth]{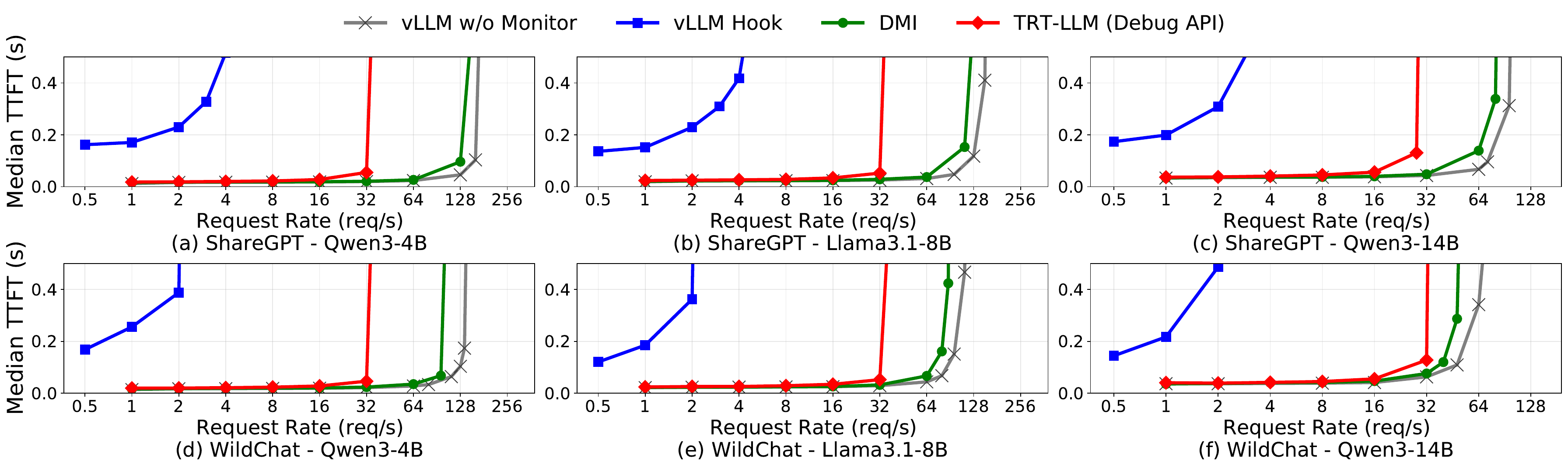}
    \caption{\textbf{Online Serving Performance: TTFT.}}
    \label{fig:online}
\end{figure}

\begin{figure}[t]
    \centering
    \includegraphics[width=\linewidth]{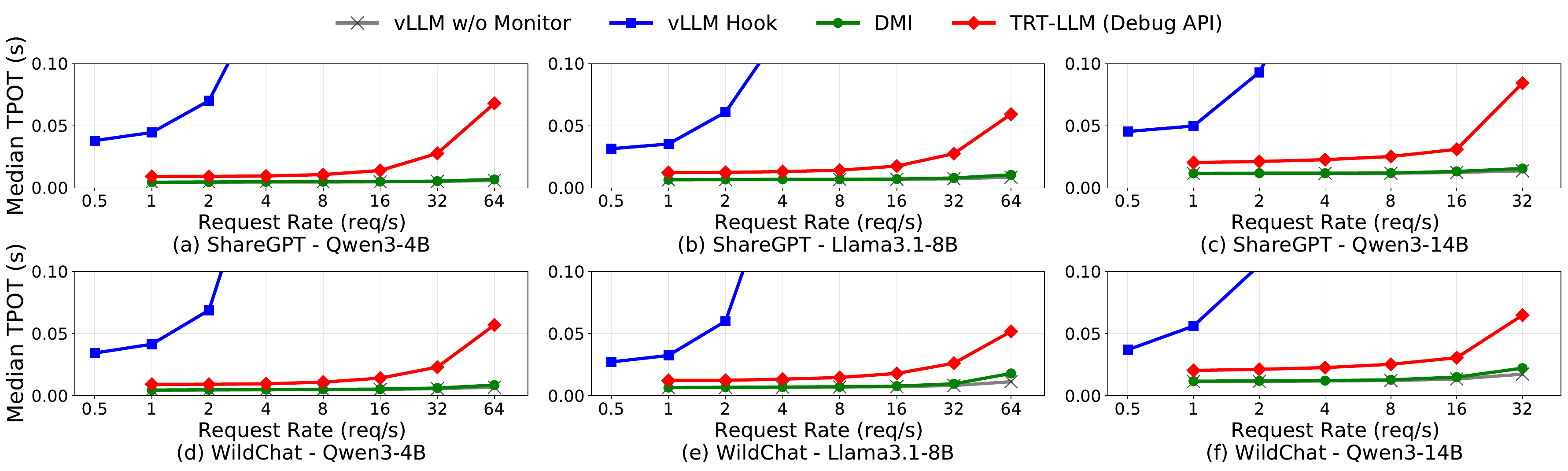}
    \caption{\textbf{Online Serving Performance: TPOT.}}
    \label{fig:online_tpot}
\end{figure}

\paragraph{Custom hooks.}
We next evaluate the setting where flexible, user-defined hook points are required. Since engine-coupled APIs cannot generally support these observation sites, we compare \NAME against the flexible hook-based baselines listed in Table~\ref{tab:baseline_hooks}. The results are shown in Figure~\ref{fig:offline_hooks}. In this setting, \NAME continues to maintain near-zero performance overhead while supporting substantially more flexible observation sites. Across all evaluated models and batch sizes, \NAME incurs only 0.4--6.8\% overhead, with an average of 3.6\%. In contrast, \textit{PyTorch forward hooks} incur 34.0--59.8\% overhead, with an average of 46.9\%, while \textit{NNsight} incurs 54.1--71.2\% overhead, with an average of 62.3\%. NNsight also runs out of memory for Qwen3-14B at batch size 64.

\paragraph{Multi-GPU performance.}
We evaluate \NAME under multi-GPU tensor parallelism (TP) to demonstrate that it remains compatible with distributed inference workloads. In this experiment, we synthesize 500 requests with prefill length 250 and decode length 700, and run offline inference on Qwen3-14B using Hugging Face with \texttt{torch.compile} enabled by default. Relative to the Hugging Face (ideal) configuration in Figure~\ref{fig:tpandstorage}(a), \NAME introduces only 1.72\%--3.3\% overhead while preserving near-identical throughput scaling: \NAME achieves a 1.50$\times$ speedup at TP2 and 2.22$\times$ at TP4, compared to 1.50$\times$ and 2.25$\times$ for the Hugging Face baseline.

\paragraph{Storage backend performance.}
We further study the end-to-end cost of persistent storage with a storage ablation. As shown in Figure~\ref{fig:tpandstorage}(b), we experiment on Qwen3-4B, using 10 batches of 64 synthetic requests (prefill=128, decode=500) under the hidden\_states+logits setting. \NAME without DB completes the full run in 384.1s. \NAME with ClickHouse on disk increases the total completion time to 402.7s, corresponding to an end-to-end overhead of 4.86\%. Placing ClickHouse in memory reduces the completion time to 389.1s, leaving only 1.31\% overhead over vanilla \NAME. The small overhead is attributable to the pipelined Data Exporter (\S\ref{sec:backend}).

\subsection{Online Serving Performance}\label{evl:online}

We evaluate \NAME under online serving conditions using vLLM-0.17.0. We introduce TensorRT-LLM~1.2.0 as an additional baseline to evaluate the monitoring overhead of a compiled inference engine. Each experiment sends requests to an inference server following a Poisson arrival process at rates from 1 to 256~req/s for 30 seconds after 50 warm-up requests. We report median TTFT (time-to-first-token) and TPOT (time-per-output-token) averaged across three randomly sampled datasets. All baselines are configured to extract all per-layer hidden states and offload them to host memory.

Figures~\ref{fig:online} and~\ref{fig:online_tpot} show the results. vLLM-Hook is unusable even at minimal load: at 1~req/s its median TTFT is already 10--15$\times$ the baseline (152--252~ms vs.\ 12--34~ms) and its TPOT is 5--8$\times$ higher, because each hook performs a synchronous D2H copy that serializes with inference and prevents CUDA graph replay. TRT-LLM (Debug API) incurs a persistent TPOT overhead at low request rates, roughly 2$\times$ the vLLM baseline (e.g., 9.3 vs.\ 4.5~ms for Qwen3-4B at 1~req/s), reflecting the cost of synchronous per-step D2H of all hidden-state tensors within the engine execution loop. At higher rates, the reduced effective throughput causes TRT-LLM's TTFT to degrade sharply far earlier than \NAME and the vLLM baseline. In contrast, \NAME adds less than 10\% TPOT overhead up to 16~req/s across all models (e.g., 7.3 vs.\ 7.0~ms for Llama-3.1-8B) and maintains near-baseline TTFT, because its asynchronous ring-buffer transport overlaps with GPU computation rather than blocking the inference pipeline.

\begin{figure}[t]
    \centering
    \includegraphics[width=0.6\linewidth]{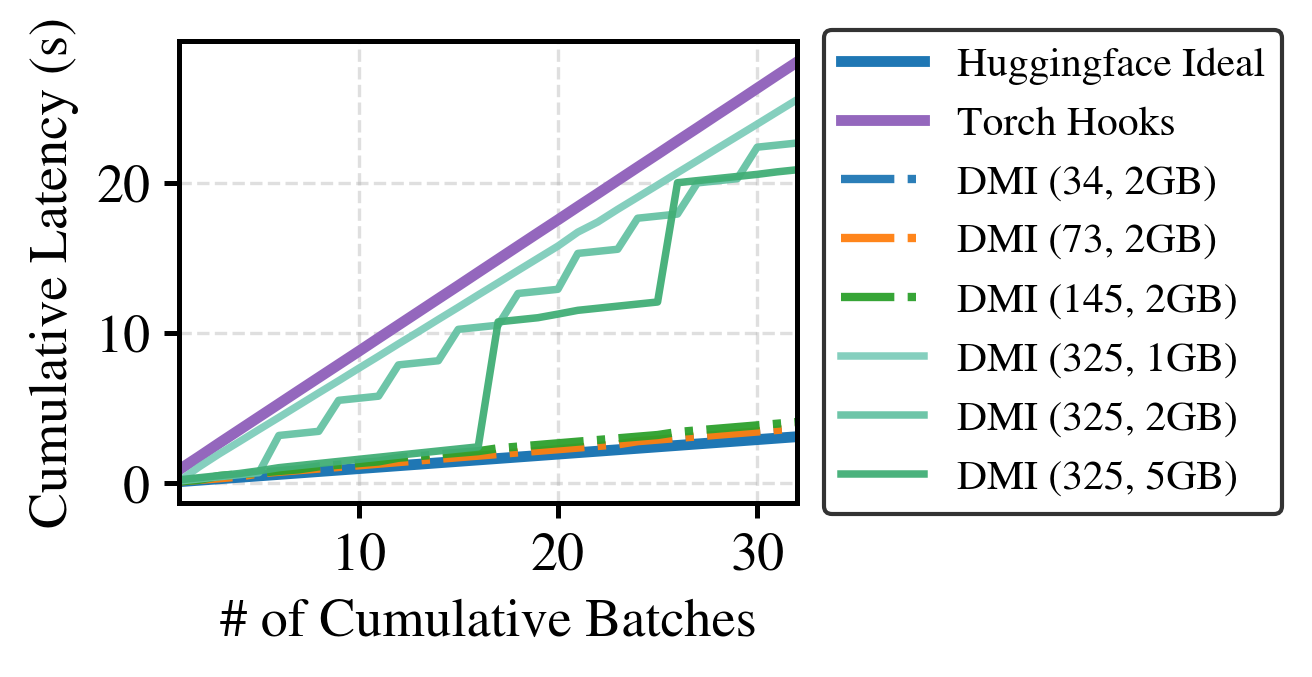}
    \caption{\textbf{\NAME behavior when generation speed of internal tensors exceeds export bandwidth.} Each \NAME (x, y) denotes x hook points and a y-sized ring buffer.}
    \label{fig:overload_onset}
\end{figure}

\begin{figure}[t]
    \centering
    \includegraphics[width=0.75\linewidth]{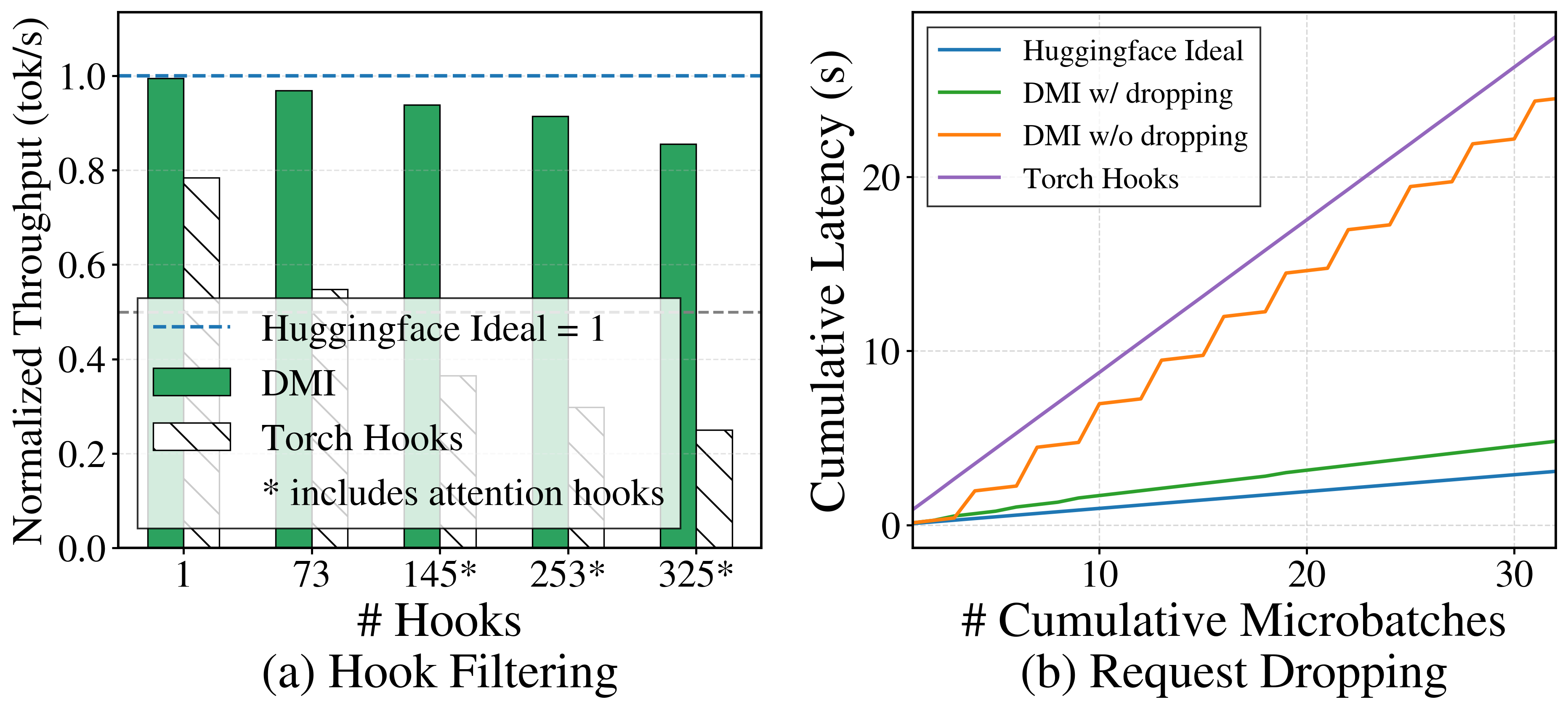}
    \caption{\textbf{Observability tradeoffs at hook and request granularity.} (a) Hook filtering trades observability for overhead by varying the enabled hook set. (b) Request-granular dropping trades request coverage for overhead by varying how many requests remain in the observation path under overload.}
    \label{fig:overload_control}
\end{figure}

\subsection{User Policy to Manage Overload}
\label{evl:overload}

We quantify \NAME's on-policy behavior under overload, where data generation approaches or exceeds sustainable PCIe export bandwidth, and expose two user policies that trade observability for usability. In a prefill-only setting, where intermediate tensors are produced fastest, Figure~\ref{fig:overload_onset} shows that a small hook set stays within the PCIe budget and incurs only 12.9\%--31.6\% overhead, whereas enabling all 325 hooks (including attention patterns from eager attention backend) drives the system into sustained overload with 574.9\%--725.9\% overhead even as larger rings merely delay this transition; \NAME still outperforms PyTorch hooks at 807.0\% overhead in this regime. Figure~\ref{fig:overload_control}(a) shows that hook filtering reduces observation volume so \NAME's overhead remains between 0.4\% and 16.9\%, while PyTorch hooks grow from 27.6\% to 300.2\%. Figure~\ref{fig:overload_control}(b) shows that \NAME's runtime policy keeps the tensor stream within sustainable PCIe bandwidth by discarding observations from a subset of requests, cutting end-to-end overhead from 692.0\% to 55.8\% without affecting inference.

\begin{figure}[t]
    \centering
    \begin{minipage}[t]{0.49\linewidth}
        \centering
        \includegraphics[width=\linewidth]{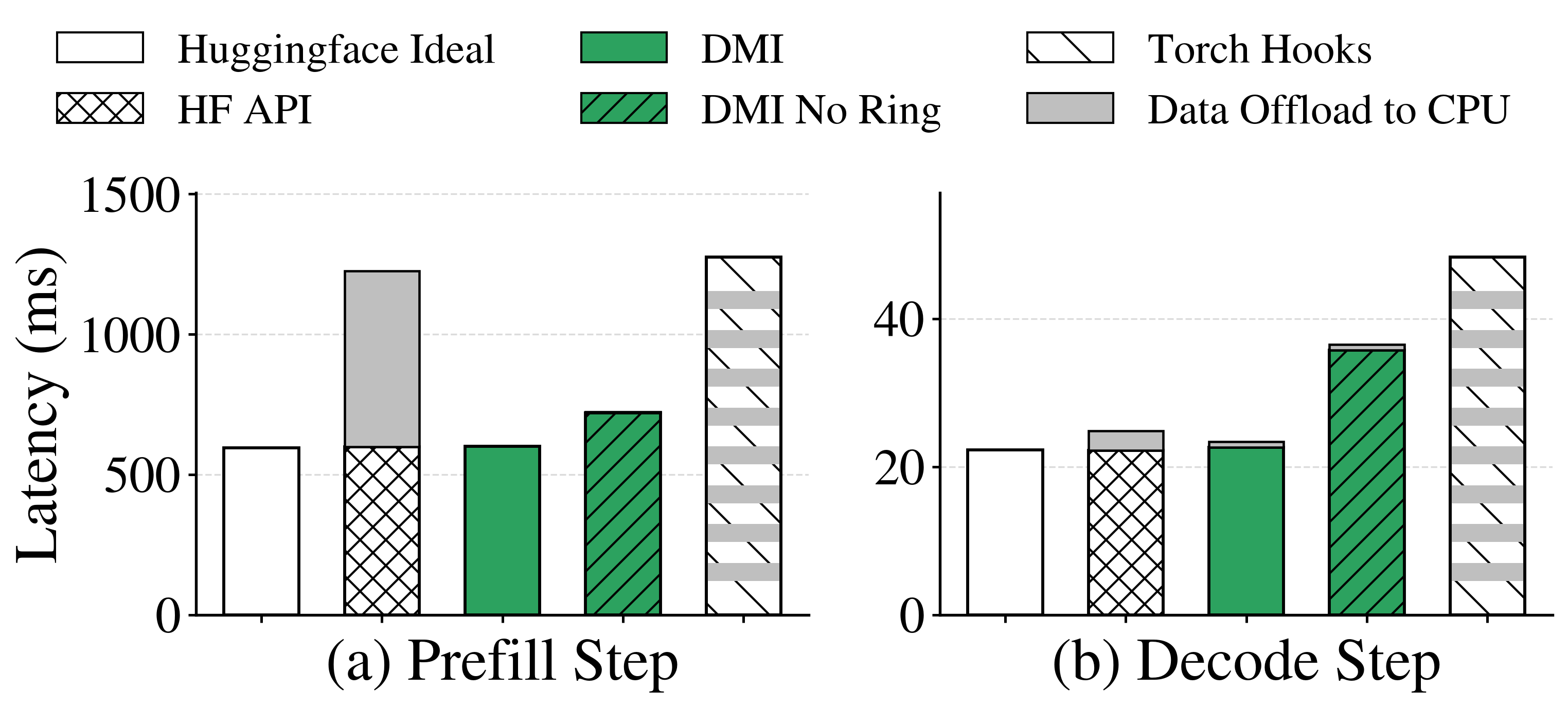}
        \caption{\textbf{Per-step overhead breakdown and ablation study.}}
        \label{fig:one_step}
    \end{minipage}
    \hfill
    \begin{minipage}[t]{0.49\linewidth}
        \centering
        \includegraphics[width=\linewidth]{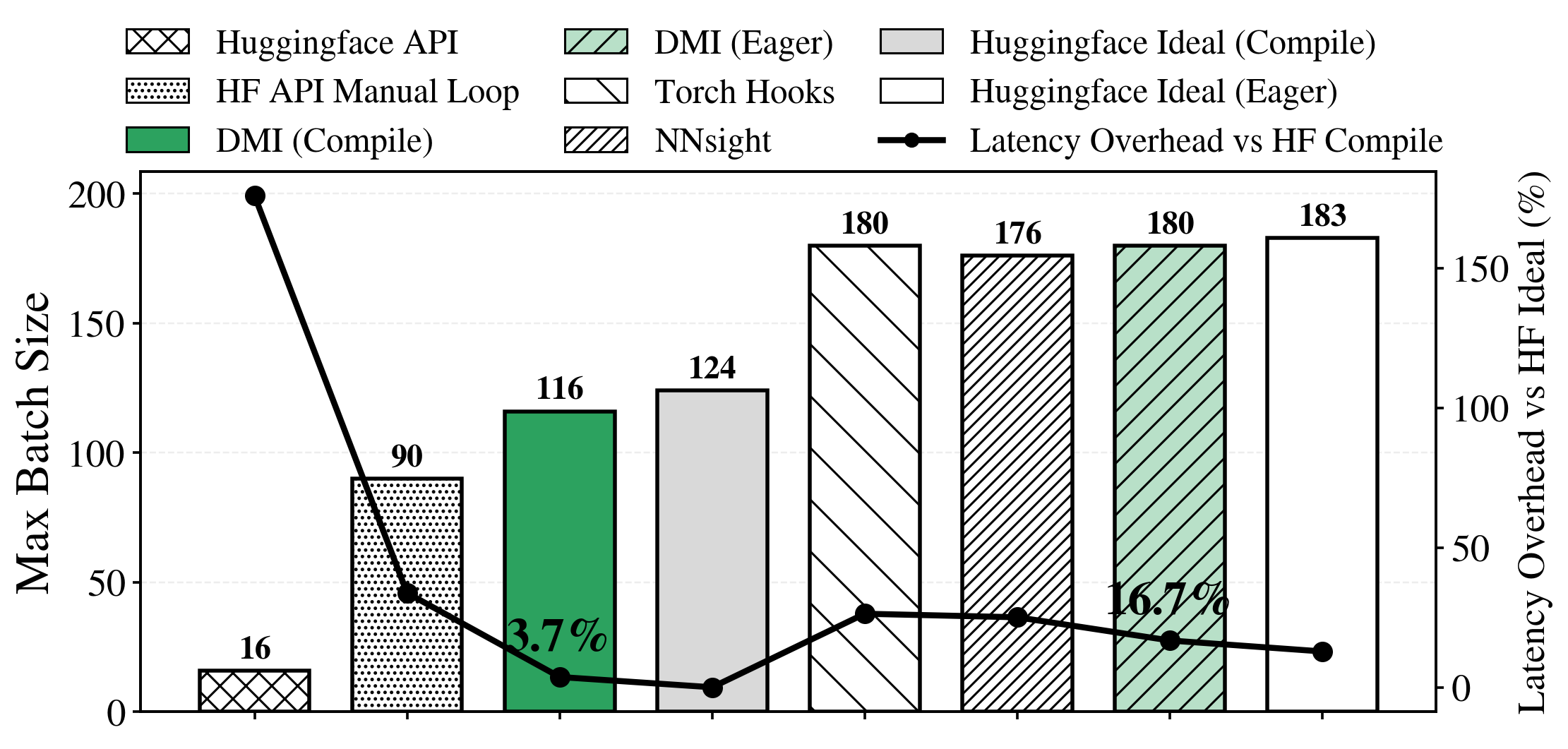}
        \caption{\textbf{Effective memory footprint and latency overhead under hidden-state and logits extraction on Qwen3-14B.}}
        \label{fig:batch_size}
    \end{minipage}
\end{figure}

\subsection{Microbenchmarks}
\label{evl:micro}

\paragraph{Per-step overhead breakdown.}
We first quantify \NAME's impact on a single inference step and break the total overhead into compute and transfer components, in order to isolate the contributions of GPU-side extraction and asynchronous offloading. In this experiment, we use Qwen3-1.7B with Hugging Face as the backend, and measure one prefill step with batch size 64 and sequence length 128, followed by ten decode steps. For decode, we report the last step, including the tail of the offload path. We run each configuration five times and report the median. The baselines are: (1) native Hugging Face, (2) Hugging Face with built-in hidden-state and logits return, (3) full \NAME with $Ring^2$, (4) \NAME without $Ring^2$, which falls back to Python-callback-style HookPoints, and (5) PyTorch forward hooks. We include the \NAME ablation with and without $Ring^2$ to directly measure the contribution of our GPU-side abstraction.

As shown in Figure~\ref{fig:one_step}, full \NAME incurs only 1.36\% overhead on prefill and 5.05\% on decode. In prefill, compute accounts for 70.9\% of the total overhead, while transfer contributes only 29.1\%, indicating that most export cost is hidden by overlap with computation. In decode, the split is 32.9\% compute and 67.1\% transfer. These results show that \NAME keeps the compute path close to native execution while substantially amortizing export cost through asynchronous draining. Without $Ring^2$, \NAME incurs 21.63\% overhead on prefill and 63.67\% on decode. In this setting, almost all of the overhead comes from computation itself: 97.8\% of the prefill overhead and 94.7\% of the decode overhead are compute-side. This is because, without GPU-side extraction and staging, observation inevitably perturbs the optimized execution path and loses the benefit of CUDA Graph replay. This ablation directly shows the importance of our GPU-side abstraction.

The other baselines illustrate the same tradeoff from different directions. Hugging Face's built-in API stays close to native execution on the compute path, but because offload is synchronous, it blocks subsequent computation and incurs 106.2\% overhead on prefill and 11.3\% on decode. PyTorch forward hooks perform worst: they both perturb the execution graph and introduce synchronous transfer, resulting in 114.4\% overhead on prefill and 116.7\% on decode. This breakdown shows that $Ring^2$ preserves the optimized compute path, while asynchronous draining prevents export from dominating per-step costs.

\paragraph{Effective memory footprint.}
We next evaluate the effectiveness of \NAME's GPU-side memory design. In this experiment, we serve requests with prefill length 250 and decode length 1000, and measure the largest batch size that can run without out-of-memory on an 80GB H100 while serving Qwen3-14B with Hugging Face. We also record the corresponding step latency to compute end-to-end throughput. All baselines extract hidden states and logits, and \NAME uses a 2GB $Ring^2$ size.

Figure~\ref{fig:batch_size} reports the results. Since eager execution and \texttt{torch.compile} have different baseline memory requirements, we compare each \NAME configuration against its corresponding Hugging Face ideal. \NAME remains the closest to the ideal in both modes: \NAME (Compile) supports a batch size of 116 vs.\ 124, and \NAME (Eager) supports 180 vs.\ 183, with only 3.7\% and 3.5\% latency overhead, respectively. Torch hooks and NNsight have similarly small memory footprints but incur higher end-to-end costs.
Hugging Face's built-in extraction APIs, both \texttt{generate()} and the manual-loop form, incur noticeably larger memory and runtime costs. Overall, \NAME's isolated GPU-side memory design preserves serving capacity while introducing minimal overhead.

\section{Use Case Demonstration}

\subsection{Mechanistic Interpretability.}
Mechanistic interpretability is fundamentally an observability task: explaining a language model's predictions requires direct inspection of internal states such as residual streams~\cite{zou2023representation} and attention patterns~\cite{clark2019bert}. Recent work has identified recurring structural patterns in attention heads associated with specific functional roles, including the \emph{Duplicate Token} pattern~\cite{wang2023ioi} and the \emph{attention-sink} pattern~\cite{xiao2023streaming}. We show that two such patterns can be observed in heads of an off-the-shelf model from a single \NAME capture.

We capture activations on a canonical Indirect Object Identification probe~\cite{wang2023ioi}, the prompt \emph{``While John and Mary were working at the office, John gave a notebook to''}, on which the model correctly continues with \emph{``Mary''}. We use Qwen3-0.6B (28 layers, 16 attention heads per layer) with eager attention so per-(query, key) attention weights are materialized. \NAME captures the full per-(layer, head, query, key) attention tensor in a single forward pass through its asynchronous $Ring^2$ export path into ClickHouse. Inspecting the captured matrices offline for recurring structural patterns, we find two heads whose attention aligns with signatures previously characterized in the literature: one matching the Duplicate Token pattern~\cite{wang2023ioi}, in which a query at a repeated token attends back to its first occurrence, and one matching the attention-sink pattern~\cite{xiao2023streaming}, in which every query routes its mass to the first position.

\begin{figure}[t]
    \centering
    \includegraphics[width=0.75\linewidth]{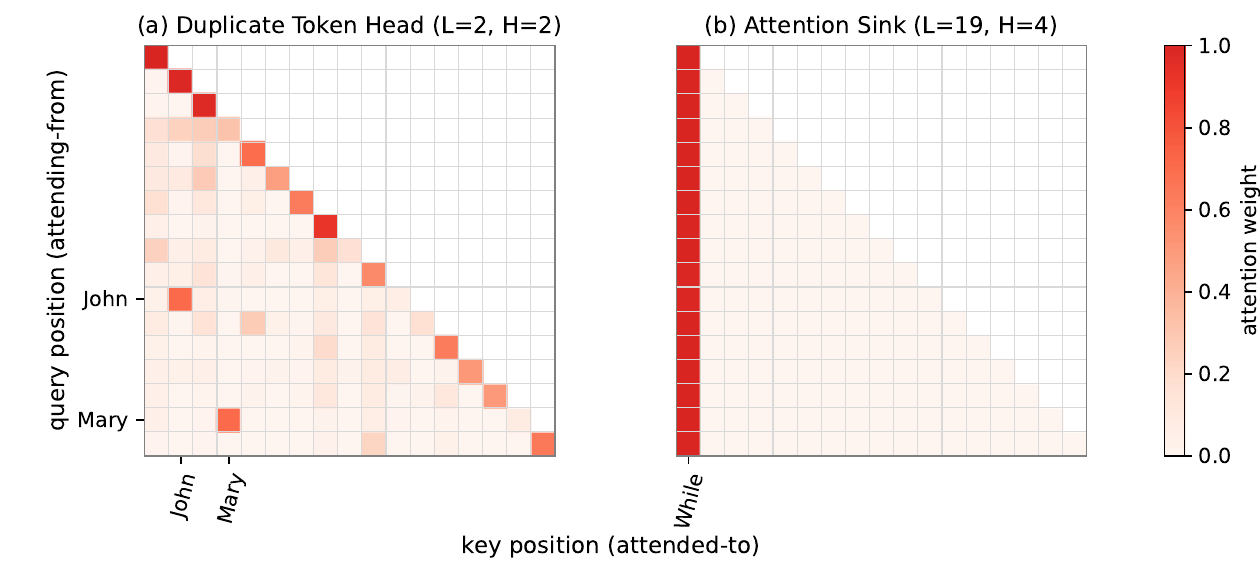}
    \caption{\textbf{Two attention-head patterns observed in Qwen3-0.6B from a single \NAME capture (17-token IOI prompt).} (a) A head at layer 2, head 2 exhibiting the \emph{Duplicate Token} pattern from~\cite{wang2023ioi}: each query at a repeated token (\emph{John}, \emph{Mary}) attends back to that token's first occurrence (corresponding column). (b) A head at layer 19, head 4 exhibiting the \emph{attention-sink} pattern from~\cite{xiao2023streaming}: every query routes its mass to the first content token \emph{While}, regardless of position.}
    \label{fig:usecase_interp_heads}
\end{figure}

Figure~\ref{fig:usecase_interp_heads} visualizes these two heads. Conventional approaches to obtaining this data require the analyst to assemble the capture, transport, and persistence pipeline themselves: ad-hoc PyTorch hooks introduce host-side overhead and break compiler-driven execution graphs, while engine-bound interfaces expose only a fixed subset of internal states. \NAME removes these costs as a single integrated package: its \texttt{HookPoint} primitive captures attention along with any other PyTorch-visible tensor, while hook selection makes it trivial to reconfigure which tensors are captured for any new analysis without writing extraction code.

\subsection{Knowledge Distillation.}
Intermediate hidden states captured by \NAME are not only observable but practically useful for downstream tasks. We demonstrate this through knowledge distillation, where a smaller student model is trained to approximate a larger teacher. Beyond matching the teacher's output logits, the student can also receive layer-wise supervision from the teacher's intermediate hidden states. We show that this additional supervision consistently improves distillation quality, and that \NAME's efficient storage makes it practical to collect these states once and reuse them across multiple student configurations.

We evaluate on HellaSwag\footnote{We use HellaSwag's standard length-normalized accuracy metric.}, a 4-way multiple-choice commonsense reasoning benchmark. We use Qwen3-8B (36 layers) as the teacher. Since the base model has not been trained on the downstream task, we split the HellaSwag dataset in half for training and evaluation, and supervised fine-tune the base model on the training portion, achieving 76.95\% accuracy. Student models of 9, 18, and 24 layers are created by copying selected layers from the fine-tuned teacher, preserving the original hidden dimension. We compare four training regimes, all using the same training portion: (1)~distillation with forward layer matching, which combines a KL divergence loss on logits with an L2-normalized MSE loss on matched hidden states; (2)~logits-only distillation, which uses only the KL loss without hidden state supervision; (3)~SFT applied to the pruned student without any teacher signal; and (4)~the pruned teacher evaluated directly without further training.

Table~\ref{tab:distill} shows the results. Distillation with hidden state matching consistently outperforms logits-only distillation by 2--3\% across all student sizes, while both methods substantially exceed SFT without a teacher by 4--9\%. This confirms that intermediate hidden states carry supervision signal beyond what output logits alone provide. Conventional distillation requires re-running the teacher for every student configuration, and this cost grows linearly with the number of student sizes and strategies explored. \NAME removes this repeated cost: its asynchronous export path can efficiently capture and persist the teacher's hidden states in a single pass, producing a reusable dataset for all subsequent student training runs without the teacher.

\begin{table}[t]
\centering
\caption{Distillation accuracy (\%) on HellaSwag. The teacher is Qwen3-8B (36 layers), supervised fine-tuned on the training portion. Students are pruned to fewer layers. Forward Matching uses hidden state MSE and logits KL; KL Only uses logits KL alone.}
\label{tab:distill}
\small
\begin{tabular}{lccc}
\toprule
\textbf{Method} & \textbf{9 Layers} & \textbf{18 Layers} & \textbf{24 Layers} \\
\midrule
Teacher (36 Layers) & \multicolumn{3}{c}{76.95} \\
\midrule
Forward Matching + KL & \textbf{36.18} & \textbf{55.14} & \textbf{64.75} \\
KL Only & 33.75 & 52.15 & 62.79 \\
SFT (no teacher) & 29.53 & 42.99 & 57.20 \\
Pruned (no training) & 26.02 & 26.46 & 41.80 \\
\bottomrule
\end{tabular}
\end{table}
\section{Related Work}
\label{sec:related}

\paragraph{Inspecting and intervening on model internals.}
Prior work on accessing model internals is dominated by model-level extension tools. PyTorch~\cite{paszke2019pytorch} exposes forward hooks to capture inputs and outputs from \texttt{nn.Module}. Captum~\cite{kokhlikyan2020captum} provides a unified interpretability library for PyTorch, while TorchLens~\cite{TorchLens}, NNsight/NDIF~\cite{fiottokaufman2024nnsightndifdemocratizingaccess}, and FlexModel~\cite{choi2023flexmodel} offer increasingly flexible access to model internals for analysis-oriented workflows. TransformerLens~\cite{nanda2022transformerlens} claims full internal access. It implements a custom transformer model to apply its hooks, rendering it unsuitable for production inference.
Different from these tools, \NAME distinguishes itself by targeting performance as a first-tier goal while retaining the highest level of internal accessibility among existing works.

\paragraph{Engine-coupled extraction and debugging interfaces.}
Existing support for model internals within inference engines is either predefined or tightly coupled to specific runtimes. Hugging Face Transformers~\cite{wolf-etal-2020-transformers} can return hidden states, attentions, and logits via generation outputs. vLLM's~\cite{vllm} recent v0.17.0 update introduced a feature to expose auxiliary hidden states for EAGLE-3~\cite{li2025eagle}; however, our investigation reveals this functionality is limited to the prefill stage and is strictly coupled with the speculative decoding framework. Similarly, the vLLM Hook library~\cite{ko2026vllm} extracts internal states but supports only a limited number of hooks. While TensorRT-LLM~\cite{trtllm} provides a debug feature for flexible state extraction, it is engine-specific and non-portable. These facilities are bound to specific engines and output interfaces, failing to provide a standalone, general observability path for high-rate, user-defined observation points during serving.

\paragraph{LLM serving and production observability.}
\NAME is complementary to existing research on high-performance LLM serving. Systems such as Orca~\cite{yu2022orca}, vLLM~\cite{vllm}, Sarathi-Serve~\cite{agrawal2024taming}, DistServe~\cite{zhong2024distserve}, and TensorRT-LLM~\cite{trtllm} prioritize throughput and latency via continuous batching, KV-cache management, phase disaggregation, and specialized runtimes. Meanwhile, production observability in serving systems typically centers on Prometheus-style metrics~\cite{prometheus}, such as request throughput, latency, cache utilization, and GPU telemetry. Other low-level monitoring frameworks~\cite{deng2025mycroft,deng2025minder} focus on GPU performance, networking, or collective communications. Our work bridges the gap between these system-level signals and arbitrary model-internal states by providing a standalone online observation substrate designed to complement different serving backbones.

\section{Conclusions}
\label{sec:conclusion}

This paper presents \NAME, an observation substrate for high-speed LLM inference that addresses the lack of observability of model internals during runtime. By combining a device-host memory staging mechanism $Ring^2$ and a dedicated asynchronous data exporter, \NAME keeps performance overhead under 7\% for both offline and online serving, effectively cutting latency penalties by 2$\times$ to 15$\times$ relative to baselines with similar observability features.

\bibliographystyle{plainnat}
\bibliography{refs}

\end{document}